\documentclass[]{interact}
\usepackage[flushleft]{threeparttable}
\usepackage{epstopdf}
\usepackage[caption=false]{subfig}

\usepackage{apacite}
\usepackage{natbib}
\usepackage{hyperref}
\usepackage{booktabs}
\usepackage{graphicx}
\makeatletter
\newcommand{\doi@}[1]{\href{https://doi.org/#1}{#1}}
\DeclareRobustCommand{\doi}{\hyper@normalise\doi@}
\makeatother

\usepackage{tablefootnote}

\theoremstyle{plain}

\theoremstyle{definition}

\theoremstyle{remark}

\begin{document}

\title{Investigating first-language bias in LLM-based automated essay scoring: A cross-prompt evaluation of an open-weight AI-model on TOEFL essays}

\author{
\name{John Maurice Gayed\textsuperscript{a}}
\affil{\textsuperscript{a} Global Education Center, Waseda University}
}

\maketitle
\begin{abstract}
This study examines the cross-prompt generalization and first-language (L1) scoring effects of a LoRA-adapted open-weight large language model (Gemma-3-27B-it) applied to automated essay scoring. Using the identical model and inference configuration reported in \cite{gayed2026aiawe}, which was fine-tuned on 480 argumentative essays from two prompts, we evaluate scoring accuracy on the full TOEFL11 corpus: 12,100 essays written by test-takers from 11 first-language backgrounds across eight prompts, none of which were seen during training. The model's raw scores (0.5--5.0) are mapped to the same three proficiency bands (low, medium, high) used by ETS, enabling direct comparison. The model achieved an overall band agreement of 77.79\% and a quadratic weighted kappa of 0.702, with adjacent-band agreement of 99.98\%. Accuracy was stable across all eight unseen prompts, with no advantage for prompts thematically related to the training data, indicating robust cross-prompt generalization. However, the model exhibited a systematic, L1-linked scoring offset. Within every proficiency band, essays from European-language backgrounds received consistently higher scores than essays from East-Asian-language backgrounds, a pattern not attributable to the composition of the fine-tuning data. This is the first large-scale L1 fairness analysis of a fine-tuned open-weight LLM for automated essay scoring.
\end{abstract}

\begin{keywords}
Automated Writing Evaluation; Large Language Models; L1 bias; language assessment
\end{keywords}

\section{\MakeUppercase{Introduction}}

The assessment of second-language English proficiency is a high-stakes and widely used component of academic admissions and placement worldwide \citep{ihlenfeldt2023meta}, and writing is a central part of what such assessments measure. Assessing that proficiency reliably and at scale remains a persistent challenge. Evaluating student writing requires attending to surface features such as grammar and word choice, in addition to higher-order qualities including coherence, task completion, and the quality of reasoning \citep{beaufort2008college, riemenschneider2024interplay}. For educators, the process is time-consuming, subjective, and difficult to standardize across large and diverse student populations \citep{finn2020applying, lesterhuis2022validity, li2022impact}.

Automated writing evaluation (AWE) and automated essay scoring (AES) have long been pursued as tools to address these challenges \citep{chen-2008, shermis2013handbook, ramesh2022automated}. The field has evolved from rule-based systems focused on surface-level features \citep{attali2013validity}, through machine learning approaches incorporating latent semantic analysis \citep{foltz2013implementation} and NLP-derived linguistic features shown to predict human judgments of writing quality \citep{crossley2018assessing, crossley2020linguistic}, to the recent application of large language models (LLMs). Earlier AWE systems could not reliably capture higher-order dimensions of writing such as coherence and argumentation quality \citep{ramesh2022automated}; LLMs have since demonstrated promise in evaluating precisely these dimensions \citep{wang2024effectiveness, xiao2024automation}. Fine-tuning LLMs on even modest quantities of human-scored data has been shown to produce substantial improvements in scoring accuracy over zero-shot or few-shot prompting \citep{wang2024effectiveness, xiao2024automation, latif2024finetuning}, and recent systematic reviews confirm that while performance varies across tasks and configurations, fine-tuned LLMs can approach human inter-rater reliability on several benchmarks \citep{emirtekin2025systematic, xu2024systematic}.

However, two critical dimensions of LLM-based AES remain underexplored. The first is cross-prompt generalization. Whether a model trained on essays responding to a small number of specific prompts can score essays on entirely different topics with comparable accuracy. Many published studies evaluate LLM scoring on the same prompts used for training or on closely related ones, leaving open the question of how well performance transfers to unseen writing tasks \citep{xu2024systematic}. This is an important practical question, since an operational scoring system must handle a variety of essay topics without retraining for each one.

The second underexplored dimension is first-language (L1) fairness. Whether the model's scoring accuracy varies systematically across writers from different first-language backgrounds. L2 writing is shaped by L1-specific transfer effects, including rhetorical organization patterns, error types, and lexical preferences. If an LLM scorer assigns systematically different accuracy levels to writers depending on their L1, this constitutes a form of differential item functioning that compromises the fairness of the assessment \citep{liu2025enhancing}. Despite its importance, L1 fairness has received limited attention in the LLM-based AES literature. \cite{liu2025enhancing} examined L1-related scoring effects in a fine-tuned GPT model across four L1 groups, finding that the model was systematically more lenient than human raters, most markedly for German L1 writers, and \cite{gayed2026aiawe} explicitly identified the absence of an L1 analysis as a limitation of that study.

The present study addresses both gaps. We take the identical LoRA-adapted Gemma-3-27B-it model reported by \cite{gayed2026aiawe}, which was fine-tuned on the full 480-essay ETS dataset from two prompts. In \cite{gayed2026aiawe}, the same model architecture and LoRA configuration achieved a quadratic weighted kappa (QWK) of 0.828 and an agreement rate of 90.56\% within $\pm$0.5 of the human score on a 360-essay held-out set (using a 120-essay training split); the production adapter used here was trained on all 480 essays. This configuration was selected because it was the best-performing model in that study, outperforming both the larger LLaMA-3.3-70B under identical LoRA settings and the fine-tuned GPT-3.5 baseline; the present study asks whether this best-validated configuration holds up when moved from a small, in-domain held-out set to a large corpus of unseen prompts and writers. Evaluating a single fixed model is therefore a deliberate design choice rather than a convenience. We hold the model, adapter, and inference configuration constant to ensure any effects observed here reflect the deployed system's behavior on out-of-distribution data, not adjustments made to fit the new corpus. Without any modification to the model, its adapter, or its inference configuration, we evaluate it on the TOEFL11 corpus \citep{blanchard2013toefl11}: 12,100 essays written by test-takers from 11 L1 backgrounds across eight argumentative prompts. Crucially, none of the eight TOEFL11 prompts overlap with the two prompts used to train the model, making this a pure cross-prompt generalization test. In addition, the inclusion of 11 diverse L1 groups, with 1,100 essays per language, provides the necessary data for a systematic L1 fairness analysis.

Three research questions guide the investigation:

\begin{itemize}
    \item {RQ1:} How accurately does the LoRA-adapted Gemma-3-27B-it model classify unseen argumentative essays into proficiency bands (low, medium, high)?
    \item {RQ2:} Does the model's scoring accuracy vary across the eight essay prompts, and does thematic proximity to the training prompts influence performance?
    \item {RQ3:} Does the writer's first language influence the model's scoring accuracy, and if so, are these effects consistent across proficiency bands?
\end{itemize}


\section{\MakeUppercase{Literature Review}}

\subsection{LLM-based automated essay scoring}

The application of LLMs to essay scoring has expanded rapidly since 2023. \cite{mizumoto2023exploring} applied GPT-3 (text-davinci-003) to scoring 12,100 TOEFL essays via zero-shot prompting, achieving a QWK of 0.388 against the ETS benchmark proficiency levels, rising to 0.605 when GPT scores were combined with linguistic features. The model's scoring consistency across repeated runs was considerably higher (QWK = 0.682), indicating that the primary limitation lay in accuracy rather than stability. These results demonstrated the potential of LLMs for AES.

Subsequent work established that task-specific fine-tuning substantially improves performance over prompting alone. \cite{wang2024effectiveness} fine-tuned GPT-3.5 on the ETS TOEFL Independent Writing dataset (480 essays, two prompts) and, for the best-performing model (fine-tuned on a single prompt), achieved a QWK of 0.78 and an agreement rate of 84.72\%, substantially outperforming both zero-shot GPT-3.5 and the more capable GPT-4 under identical conditions. \cite{xiao2024automation} confirmed this pattern, demonstrating that a fine-tuned GPT-3.5 surpassed the non-fine-tuned GPT-4 on QWK, and further showed that LLM-generated feedback can enhance human grading accuracy. \cite{latif2024finetuning} demonstrated that fine-tuned GPT-3.5 outperformed fine-tuned BERT by an average of 9.1\% in scoring accuracy across six science assessment tasks.

\cite{pack2024large} evaluated four LLMs (GPT-4, GPT-3.5, PaLM 2, and Claude 2) on English language learner writing, measuring both inter-rater and intra-rater reliability. GPT-4 achieved the highest agreement with human raters, but the study also documented temporal instability: correlations between model and human scores declined between two time points, suggesting that ongoing model updates can undermine scoring consistency. \cite{tate2024can} examined GPT-3.5 and GPT-4 in a zero-shot setup across three different writing corpora, finding that while GPT-4 had better internal consistency, its agreement with human scores was moderate at best (weighted kappa of 0.58 for non-English learners), reinforcing the conclusion that fine-tuning is necessary for reliable scoring.

Two recent systematic reviews consolidate these findings. \cite{emirtekin2025systematic} analyzed 49 peer-reviewed studies on LLM-powered assessment, finding that while some configurations achieved very high agreement with human raters, performance varied considerably across tasks and contexts. \cite{xu2024systematic} concluded that AWE-based assessment cannot yet fully replace human judgment, citing limitations in accuracy, reliability, and robustness to unseen prompts.

A common thread across these studies is the reliance on proprietary APIs, most often OpenAI's GPT family. \cite{gayed2026aiawe} addressed this gap by fine-tuning open-weight models (Gemma-3-27B-it and LLaMA-3.3-70B) using Low-Rank Adaptation (LoRA) on the same 480-essay ETS dataset used by \cite{wang2024effectiveness}. The fine-tuned Gemma model achieved a QWK of 0.828 and an agreement rate of 90.56\%, outperforming both LLaMA-3.3-70B and the GPT-3.5 baseline across all metrics. That study mainly found that open-weight models can match or exceed proprietary models for rubric-aligned scoring and model scale is not a reliable predictor of performance. However, that study evaluated on only 360 essays from two prompts and did not examine L1 effects, explicitly identifying both as directions for future work.

\subsection{Cross-prompt generalization}

A key practical requirement for any deployed AES system is the ability to score essays on topics not seen during training. This cross-prompt generalization problem has been recognized in the AES literature, but relatively few LLM-based studies have tested it rigorously.

\cite{wang2024effectiveness} provided early evidence that fine-tuned GPT-3.5 could generalize across prompts, though the study evaluated this only within the two prompts used for training (training on one and testing on the other). More recently, \cite{liu2025comparing} compared six GPT-based approaches on essays drawn from the TOEFL11 corpus, reporting QWKs of 0.81 (computed on the corpus's collapsed three-band scale) for two models fine-tuned on 3,192 corpus essays. However, their evaluation employed test sets of 150 and 798 essays drawn from the same corpus, and hence the same eight prompts, as their fine-tuning data; performance was not analyzed by prompt, and their design therefore does not test generalization to unseen writing tasks.

The present study extends this line of work by introducing a stricter test of cross-prompt generalization, this study's model was trained on two prompts that are entirely outside the TOEFL11 prompt set, creating complete separation between training and evaluation topics.

\subsection{First-language effects in automated writing assessment}

It is well established that a writer's first language influences their L2 writing. L1-specific transfer effects manifest in rhetorical organization, syntactic complexity, error patterns, and lexical choice. The TOEFL11 corpus was designed in part to support research on L1-related variation in non-native English writing \citep{blanchard2013toefl11}, and the score distributions across L1 groups in the corpus reflect genuine differences in the English writing proficiency of test-taker populations from different L1 backgrounds.

Whether automated scoring systems treat essays from different L1 backgrounds equitably is a separate and important question. \cite{mizumoto2023exploring} scored the full ETS corpus of non-native English writing but did not disaggregate results by L1. \cite{liu2025enhancing} initiated the investigation of L1 effects in LLM-based scoring, evaluating a fine-tuned GPT-3.5 model on an 880-essay test set drawn from four TOEFL11 L1 groups (Chinese, French, German, and Spanish; 220 essays per group). Using the standardized mean difference between model and human ratings as a fairness index, they found that the model was more lenient than human raters for all four groups, most markedly for German L1 writers, and met the ETS reliability thresholds only for the Spanish group. Their study, however, was limited in three respects that the present study addresses. First, its scope was four of the eleven TOEFL11 L1 groups and 880 test essays, with some L1 $\times$ level cells containing as few as four essays. Second, the fine-tuned model had been trained on 2,640 essays drawn from the same corpus, the same eight prompts, and the same four L1 groups on which its fairness was then assessed, so the reported L1 effects are partly a property of the training split rather than an out-of-distribution measurement. Third, their group-level agreement comparisons were not stratified by proficiency level, a confound the authors themselves noted when attributing the low agreement for the Chinese group to its larger share of low-level essays. The present study attempts to addresses these gaps in the literature.

\section{\MakeUppercase{Methodology}}

\subsection{The scoring model}

The scoring model used in this study is identical to the one reported by \cite{gayed2026aiawe}: a Gemma-3-27B-it model \citep{gemmateam2025gemma3} adapted with a LoRA adapter \citep{hu2022lora} trained on the full 480-essay ETS TOEFL Independent Writing dataset from two prompts (the production adapter described in that paper). The LoRA adapter (\texttt{gemmalorafull}) is publicly available on the Hugging Face Hub.\footnote{\url{https://huggingface.co/jgayed/gemmalorafull}} No modifications were made to the model, adapter, or inference configuration for the present study; all hyperparameters are as reported in \cite{gayed2026aiawe} and summarized in Table~\ref{tab:model_config} for convenience.

\begin{table}[ht]
\centering
\caption{Model and inference configuration (identical to \protect\cite{gayed2026aiawe}).}
\label{tab:model_config}
\begin{tabular}{ll}
\toprule
\textbf{Parameter} & \textbf{Value} \\
\midrule
Base model & Gemma-3-27B-it \\
Adaptation method & LoRA \\
LoRA rank ($r$) & 64 \\
LoRA scaling ($\alpha$) & 128 \\
Target modules & q,k,v,o\_proj; up,down,gate\_proj \\
Training data & 480 essays, 2 prompts (full set) \\
Inference engine & llama.cpp \\
Quantization & Q4\_K\_M ($\approx$4-bit GGUF) \\
Temperature & 0 \\
\bottomrule
\end{tabular}
\end{table}

During fine-tuning, each training instance consisted of a system prompt containing the full TOEFL Independent Writing rubric (score levels 0--5 with descriptors), a user prompt containing the essay topic and the essay text, and a target output consisting of the bare human-assigned score as a decimal (e.g., 3.5). No reasoning, JSON formatting, or qualitative feedback was included in the training targets; the model was trained purely on score prediction. At inference time, the same system prompt and input format were used. Full details of the training procedure, prompts, and infrastructure are provided in \cite{gayed2026aiawe}.

The two training prompts were:
\begin{itemize}
    \item \textbf{TP-A:} ``Do you agree or disagree with the following statement? In today's world, the ability to cooperate well with others is far more important than it was in the past. Use specific reasons and examples to support your answer.''
    \item \textbf{TP-B:} ``Do you agree or disagree with the following statement? It is more important to choose to study subjects you are interested in than to choose subjects to prepare for a job or career. Use specific reasons and examples to support your answer.''
\end{itemize}

Neither of these prompts appears among the eight TOEFL11 evaluation prompts (see Appendix~\ref{appendix:prompts}), establishing complete separation between training and test data.

\subsection{Test dataset: the TOEFL11 corpus}

The TOEFL11 corpus \citep{blanchard2013toefl11} is a publicly available collection of 12,100 essays written during the TOEFL iBT test in 2006--2007. The corpus was designed to support research on native language identification, grammatical error detection, and automated essay scoring. It contains 1,100 essays from each of 11 first-language (L1) groups: Arabic (ARA), Chinese (ZHO), French (FRA), German (DEU), Hindi (HIN), Italian (ITA), Japanese (JPN), Korean (KOR), Spanish (SPA), Telugu (TEL), and Turkish (TUR). These 11 languages span seven language families: Romance (French, Italian, Spanish), Germanic (German), Indo-Iranian (Hindi), Altaic (Japanese, Korean, Turkish), Sino-Tibetan (Chinese), Afro-Asiatic (Arabic), and Dravidian (Telugu) \citep{blanchard2013toefl11}.

Essays are distributed across eight argumentative prompts (P1--P8; see Appendix~\ref{appendix:prompts}). The distribution across prompts is approximately balanced, ranging from 960 essays for P6 to 1,686 for P7. The distribution is perfectly balanced across L1 groups (1,100 per language).

Each essay was scored by at least two trained ETS raters on a 5-point scale. When scores differed by no more than one point, they were averaged and larger discrepancies triggered a third rater, with further adjudication rules as described in \cite{blanchard2013toefl11}. The resulting scores (in 0.5 increments) were then collapsed into three proficiency bands: low (1.0--2.0), medium (2.5--3.5), and high (4.0--5.0). Only these band-level scores are provided in the public corpus. The raw averaged scores are not available.

The band distribution across the full corpus is: low = 1,330 (11.0\%), medium = 6,568 (54.3\%), and high = 4,202 (34.7\%). This distribution varies substantially across L1 groups, reflecting genuine differences in the English writing proficiency of test-taker populations. 

\begin{table}[ht]
\centering
\caption{Distribution of essays by L1 and ETS proficiency band in the TOEFL11 corpus ($N$ = 12,100).}
\label{tab:corpus_l1_band}
\begin{tabular}{lccccccc}
\toprule
 & \multicolumn{3}{c}{Count} & & \multicolumn{3}{c}{Percentage} \\
\cmidrule{2-4} \cmidrule{6-8}
\textbf{L1} & \textbf{Low} & \textbf{Med} & \textbf{High} & \textbf{Total} & \textbf{Low} & \textbf{Med} & \textbf{High} \\
\midrule
Arabic   & 296 & 605 & 199 & 1,100 & 26.9 & 55.0 & 18.1 \\
Chinese  &  98 & 727 & 275 & 1,100 &  8.9 & 66.1 & 25.0 \\
French   &  63 & 577 & 460 & 1,100 &  5.7 & 52.5 & 41.8 \\
German   &  15 & 412 & 673 & 1,100 &  1.4 & 37.5 & 61.2 \\
Hindi    &  29 & 429 & 642 & 1,100 &  2.6 & 39.0 & 58.4 \\
Italian  & 164 & 623 & 313 & 1,100 & 14.9 & 56.6 & 28.5 \\
Japanese & 233 & 679 & 188 & 1,100 & 21.2 & 61.7 & 17.1 \\
Korean   & 169 & 678 & 253 & 1,100 & 15.4 & 61.6 & 23.0 \\
Spanish  &  79 & 563 & 458 & 1,100 &  7.2 & 51.2 & 41.6 \\
Telugu   &  94 & 659 & 347 & 1,100 &  8.5 & 59.9 & 31.5 \\
Turkish  &  90 & 616 & 394 & 1,100 &  8.2 & 56.0 & 35.8 \\
\midrule
\textbf{Total} & 1,330 & 6,568 & 4,202 & 12,100 & 11.0 & 54.3 & 34.7 \\
\bottomrule
\end{tabular}
\end{table}

As Table~\ref{tab:corpus_l1_band} shows, the band composition differs markedly across L1 groups. For example, 61.2\% of German L1 essays received a high band rating compared to only 17.1\% of Japanese L1 essays, while 26.9\% of Arabic L1 essays were rated low compared to only 1.4\% of German L1 essays. These distributional differences are important for the analysis. Since model accuracy may differ across bands, any comparison of overall accuracy across L1 groups is potentially confounded by band composition. For this reason, the L1 analysis in this study employs band-stratified comparisons (Section~\ref{sec:analysis_plan}).

\subsection{Scoring procedure}

Each of the 12,100 TOEFL11 essays was processed through the same inference pipeline used in \cite{gayed2026aiawe}. The system prompt contained the full TOEFL Independent Writing rubric (score levels 0--5 with descriptors; see Appendix~A.1 of \citealt{gayed2026aiawe}). The user prompt contained the relevant essay topic and the essay text. With the temperature set to 0, the model produced a deterministic raw score on the rubric's 0--5 scale in 0.5-point increments for each essay; in practice, all essays received scores between 1.0 and 5.0 except two, which received 0.5.

The model's raw scores were then mapped to the same three proficiency bands used by ETS for the TOEFL11 corpus, following the procedure described by \cite{blanchard2013toefl11}: scores of 1.0--2.0 were classified as low, scores of 2.5--3.5 as medium, and scores of 4.0--5.0 as high. This mapping enabled direct comparison between the model's band assignments and the ETS ground-truth bands.

It is important to note that the model was trained to predict raw scores (continuous values on the 0--5 scale), not band labels. The band mapping was applied post hoc for the purpose of comparison with the TOEFL11 ground truth, which provides only band-level scores. This means the model is performing a finer-grained scoring task than the evaluation can capture. In other words, two essays might both be correctly classified into the medium band, but the model's raw scores might differ substantially (e.g., 2.5 vs.\ 3.5). This asymmetry between model output and ground-truth granularity is a characteristic of the TOEFL11 corpus, not a limitation of the model.

\subsection{Analysis plan}
\label{sec:analysis_plan}

\subsubsection{Overall performance (RQ1)}

To assess overall scoring accuracy, we computed band-level exact agreement (the percentage of essays for which the model's band matched the ETS band), quadratic weighted kappa (QWK) on the three-level ordinal scale, and a confusion matrix showing the joint distribution of model and ETS band assignments. We also report the distribution of the model's raw scores to characterize its scoring behavior on the unseen TOEFL11 data.

\subsubsection{Prompt-level analysis (RQ2)}

To assess cross-prompt generalization, we computed QWK and band agreement for each of the eight TOEFL11 prompts separately. A chi-square test was used to assess whether prompt type significantly predicted model accuracy. We further examined whether prompts thematically closer to the two training prompts showed higher accuracy. While thematic proximity was assessed qualitatively rather than through a formal metric. 

\subsubsection{L1-level analysis (RQ3)}
\label{sec:l1_analysis}

The L1 analysis was conducted in three stages. First, overall accuracy and QWK were computed for each of the 11 L1 groups. Second, and critically, accuracy was computed within each proficiency band for each L1 group. This band-stratified approach is essential because, as noted above, the band composition of the TOEFL11 corpus varies substantially across L1 groups. A model that is less accurate on low-band essays will appear to perform worse overall on L1 groups with a higher proportion of low-band essays (e.g., Arabic, Japanese), even if its within-band accuracy is identical across all L1 groups. By stratifying the analysis by band, we can distinguish between genuine L1-related accuracy differences and artifacts of band composition.

Third, we examined the distribution of the model's raw scores within each ETS band, disaggregated by L1. This analysis exploits the fact that the model produces scores on a finer scale (0.5 increments) than the band-level ground truth. Even when the model correctly classifies an essay into its ETS band, the raw score assigned may vary. For example, two essays both classified as medium may receive raw scores of 2.5 and 3.5. If the model systematically assigns higher or lower raw scores to essays from certain L1 groups within the same band, this would constitute evidence of differential scoring behavior that is not visible in the band-level agreement statistics. Within-band differences were tested with chi-square tests for classification accuracy, and with Kruskal--Wallis and Mann--Whitney tests for raw-score distributions; Spearman rank correlations were used to assess the consistency of the L1 ordering across bands. An exploratory analysis of L1 patterns by language region (European, East-Asian, other) was also conducted.

\section{\MakeUppercase{Results}}

\subsection{Overall performance (RQ1)}
\label{sec:overall_performance}

Overall, the model's band assignments matched the ETS ground-truth bands in 77.79\% of cases (9,413 of 12,100), with a quadratic weighted kappa (QWK) of 0.702 (Table~\ref{tab:overall}). Disagreements were almost entirely confined to adjacent bands: adjacent agreement (model and ETS bands differing by no more than one level) reached 99.98\%, meaning that only 2 of the 12,100 essays were misclassified by two full bands (i.e., low versus high). This indicates that, while the model does not reproduce the ETS bands exactly, its errors are small in magnitude and it virtually never confuses the extremes of the proficiency scale.

\begin{table}[ht]
\centering
\caption{Overall scoring performance: Gemma band assignments vs.\ ETS proficiency bands ($N$ = 12,100).}
\label{tab:overall}
\begin{tabular}{ll}
\toprule
\textbf{Metric} & \textbf{Value} \\
\midrule
Exact band agreement & 77.79\% \\
Quadratic weighted kappa (QWK) & 0.702 \\
Adjacent band agreement (within one band) & 99.98\% \\
\bottomrule
\end{tabular}
\end{table}

The confusion matrix (Table~\ref{tab:confusion}) reveals that accuracy was not uniform across the three bands. The model was most accurate on medium-band essays (84.9\% correctly classified), followed by high-band essays (73.0\%), and was least accurate on low-band essays (57.6\%). The dominant error pattern was upward, 42.3\% of essays rated low by ETS were classified as medium by the model. In addition 27.0\% of high-band essays were classified as medium. The model showed a tendency to compress scores toward the middle of the scale, assigning the medium band to 60.1\% of all essays compared to the ETS distribution of 54.3\%, while assigning the low band less often than ETS (8.5\% vs.\ 11.0\%).

\begin{table}[ht]
\centering
\caption{Confusion matrix: ETS proficiency band (rows) vs.\ Gemma band (columns). Cell percentages are row-wise (proportion of each ETS band assigned to each Gemma band).}
\label{tab:confusion}
\begin{tabular}{lcccc}
\toprule
 & \multicolumn{3}{c}{Gemma band} & \\
\cmidrule{2-4}
\textbf{ETS band} & \textbf{Low} & \textbf{Medium} & \textbf{High} & \textbf{Total} \\
\midrule
Low    & 766 (57.6\%) & 562 (42.3\%) & 2 (0.2\%) & 1,330 \\
Medium & 268 (4.1\%) & 5,579 (84.9\%) & 721 (11.0\%) & 6,568 \\
High   & 0 (0.0\%) & 1,134 (27.0\%) & 3,068 (73.0\%) & 4,202 \\
\midrule
\textbf{Total} & 1,034 & 7,275 & 3,791 & 12,100 \\
\bottomrule
\end{tabular}
\end{table}

The distribution of the model's underlying raw scores (Table~\ref{tab:rawdist}) shows that it used almost the entire 0--5 scale (observed scores ranged from 0.5 to 5.0), with a modal score of 3.0 (22.6\% of essays) and a concentration of scores in the 2.5--4.0 range (78.6\% of all essays). Only two essays received a score below 1.0 (0.5), and these were retained in the low band for analysis. The concentration of raw scores around the middle of the scale is consistent with the upward compression of low-band essays observed in the confusion matrix.

\begin{table}[ht]
\centering
\caption{Distribution of Gemma raw scores across all essays ($N$ = 12,100).}
\label{tab:rawdist}
\begin{tabular}{lcccccccccc}
\toprule
\textbf{Score} & 0.5 & 1.0 & 1.5 & 2.0 & 2.5 & 3.0 & 3.5 & 4.0 & 4.5 & 5.0 \\
\midrule
$n$ & 2 & 94 & 204 & 734 & 1,989 & 2,733 & 2,553 & 2,240 & 1,199 & 352 \\
\% & 0.0 & 0.8 & 1.7 & 6.1 & 16.4 & 22.6 & 21.1 & 18.5 & 9.9 & 2.9 \\
\bottomrule
\end{tabular}
\end{table}

\subsection{Cross-prompt performance (RQ2)}

Performance was stable across the eight prompts, none of which appeared in the model's training data. QWK ranged narrowly from 0.677 (P8) to 0.730 (P5), and exact agreement ranged from 76.8\% (P8) to 80.5\% (P5) (Table~\ref{tab:by_prompt}). A chi-square test of independence found no significant association between prompt and the probability of correct classification, $\chi^2(7, N = 12{,}100) = 8.51$, $p = .290$, with a negligible effect size (Cramer's $V = 0.03$).

\begin{table}[ht]
\centering
\caption{Model performance by essay prompt. Accuracy within band is the proportion of essays in each ETS band correctly classified.}
\label{tab:by_prompt}
\begin{tabular}{lcccccc}
\toprule
 & & & & \multicolumn{3}{c}{Accuracy within ETS band} \\
\cmidrule{5-7}
\textbf{Prompt} & $n$ & \textbf{QWK} & \textbf{Agreement} & \textbf{Low} & \textbf{Med} & \textbf{High} \\
\midrule
P1 & 1,656 & 0.707 & 77.8\% & 54.5\% & 81.6\% & 78.6\% \\
P2 & 1,562 & 0.681 & 77.6\% & 58.8\% & 82.5\% & 74.6\% \\
P3 & 1,396 & 0.724 & 77.1\% & 62.1\% & 83.3\% & 74.0\% \\
P4 & 1,509 & 0.707 & 77.6\% & 56.4\% & 86.5\% & 71.2\% \\
P5 & 1,648 & 0.730 & 80.5\% & 58.5\% & 87.7\% & 75.9\% \\
P6 & 960 & 0.681 & 77.3\% & 60.2\% & 89.5\% & 63.4\% \\
P7 & 1,686 & 0.696 & 77.4\% & 58.2\% & 84.1\% & 72.6\% \\
P8 & 1,683 & 0.677 & 76.8\% & 53.5\% & 85.9\% & 69.0\% \\
\bottomrule
\end{tabular}
\end{table}

We further examined whether prompts thematically closer to the two training prompts were scored more accurately. Subjectively, we note that training prompt TP-B (choosing subjects by interest vs.\ career) is thematically adjacent to test prompt P1 (broad knowledge vs.\ specialization), and training prompt TP-A (cooperation) shares conceptual ground with P3 (young people helping communities). Grouping the two test prompts judged thematically adjacent to the training prompts and comparing them to the remaining six prompts revealed no advantage. The thematically close prompts yielded QWK = 0.716 and 77.5\% agreement, essentially identical to the thematically distant prompts (QWK = 0.698, 77.9\% agreement). The model's scoring accuracy was therefore not measurably influenced by thematic proximity to its training data.

\subsection{L1 effects on scoring accuracy (RQ3)}

\subsubsection{Band-stratified accuracy}

Overall QWK by L1 ranged from 0.531 (German) to 0.706 (Turkish), and overall agreement ranged from 75.6\% (Italian) to 79.8\% (Turkish) (Table~\ref{tab:l1_band_accuracy}). However, these overall figures are confounded by the differing band composition of each L1 group (Table~\ref{tab:corpus_l1_band}). Since the model is least accurate on low-band essays and German's sample is only 1.4\% low-band but 61.2\% high-band, German's relatively low overall QWK partly reflects the difficulty of the high band rather than poor performance on German writing specifically. The band-stratified accuracy figures are therefore more informative.

Within each band, accuracy varied significantly by L1 (low: $\chi^2(10) = 52.6$, $p < .001$; medium: $\chi^2(10) = 256.7$, $p < .001$; high: $\chi^2(10) = 210.5$, $p < .001$), although effect sizes were modest (Cram\'er's $V$ = 0.20, 0.20, and 0.22 respectively). The within-band accuracy spread across L1 groups was substantial. In the medium band it ranged from 63.6\% (German) to 91.9\% (Japanese), and in the high band from 51.6\% (Japanese) to 85.6\% (German). Notably, German and Japanese occupy opposite ends in both bands, but in opposite directions, a pattern explained by the analysis of raw scores below.

\begin{table}[ht]
\centering
\caption{Band-stratified accuracy by L1: proportion of essays correctly classified within each ETS proficiency band. $n$ values for each band are shown in parentheses.}
\label{tab:l1_band_accuracy}
\begin{tabular}{lccccc}
\toprule
 & & & \multicolumn{3}{c}{Accuracy within ETS band (band $n$)} \\
\cmidrule{4-6}
\textbf{L1} & \textbf{Overall} & \textbf{QWK} & \textbf{Low} & \textbf{Med} & \textbf{High} \\
 & \textbf{agree.} & & & & \\
\midrule
Arabic   & 76.1\% & 0.687 & 59.5\% (296) & 89.8\% (605) & 59.3\% (199) \\
Chinese  & 78.5\% & 0.625 & 57.1\% (98) & 89.4\% (727) & 57.5\% (275) \\
French   & 75.9\% & 0.628 & 46.0\% (63) & 76.4\% (577) & 79.3\% (460) \\
German   & 76.5\% & 0.531 & 20.0\% (15) & 63.6\% (412) & 85.6\% (673) \\
Hindi    & 78.4\% & 0.629 & 34.5\% (29) & 79.3\% (429) & 79.8\% (642) \\
Italian  & 75.6\% & 0.681 & 53.7\% (164) & 83.9\% (623) & 70.6\% (313) \\
Japanese & 79.1\% & 0.679 & 63.9\% (233) & 91.9\% (679) & 51.6\% (188) \\
Korean   & 79.2\% & 0.703 & 74.0\% (169) & 88.5\% (678) & 57.7\% (253) \\
Spanish  & 77.5\% & 0.671 & 44.3\% (79) & 84.2\% (563) & 75.1\% (458) \\
Telugu   & 79.1\% & 0.671 & 46.8\% (94) & 89.2\% (659) & 68.6\% (347) \\
Turkish  & 79.8\% & 0.706 & 56.7\% (90) & 86.7\% (616) & 74.4\% (394) \\
\bottomrule
\end{tabular}
\begin{tablenotes}
\small
\item \textit{Note.} Low-band $n$ varies substantially by L1: German ($n$ = 15), Hindi ($n$ = 29), and French ($n$ = 63) have far fewer low-band essays than Arabic ($n$ = 296) or Japanese ($n$ = 233). Accuracy estimates for the smallest cells (notably German low) should be interpreted with caution.
\end{tablenotes}
\end{table}

\begin{figure}[ht]
\centering
\includegraphics[width=\linewidth]{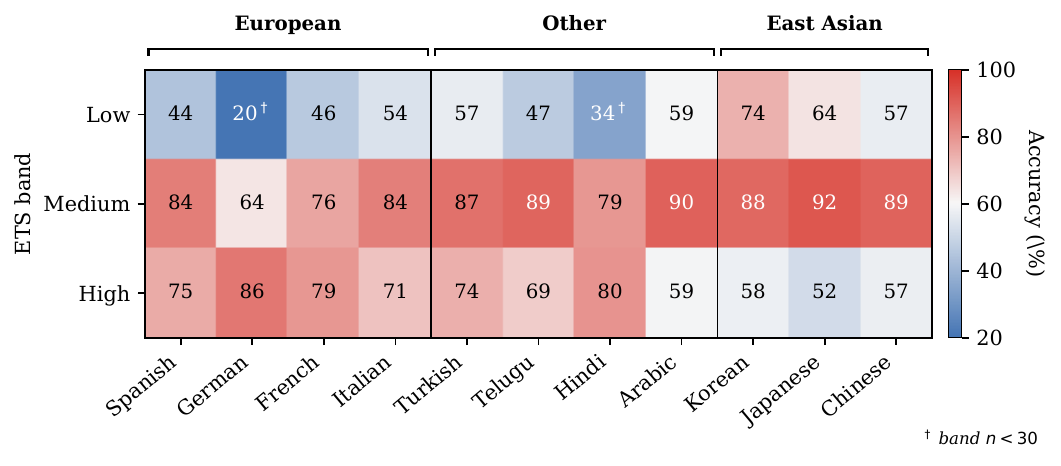}
\caption{Heatmap of classification accuracy by first language and ETS band. Each cell shows the percentage of essays from one L1 group correctly classified within one band; darker red indicates higher accuracy, darker blue lower. L1 groups are arranged in three regional clusters (European, Other, East Asian); within each cluster, languages are ordered by overall agreement rate. Daggers mark cells whose band $n$ is below 30, where estimates are less stable. The regional structure of accuracy is visible directly: European-language groups show their lowest accuracy in the low band and highest in the high band, whereas the East-Asian-language groups show the reverse pattern, consistent with a uniform downward scoring offset for East-Asian writers (see Figure~\ref{fig:l1_offset}).}
\label{fig:band_accuracy}
\end{figure}

The band-stratified accuracy is shown graphically in Figure~\ref{fig:band_accuracy}, which makes the cross-band pattern apparent: L1 groups differ markedly in which band the model scores most accurately, with some groups (e.g., Japanese, Korean) most accurate on lower bands and others (e.g., German, Hindi) most accurate on the high band.

\subsubsection{A consistent L1-linked scoring offset}

The accuracy patterns above are driven by a systematic difference in the raw scores the model assigns to essays from different L1 backgrounds. Within each ETS band, where the true proficiency level is held approximately constant, the model assigned reliably different mean raw scores depending on the writer's L1 (Kruskal--Wallis tests, all $p < .001$). Crucially, the ordering of L1 groups was highly consistent across the three bands: the rank correlation of L1 mean scores between bands was strong and positive (low vs.\ medium $\rho = 0.90$; low vs.\ high $\rho = 0.88$; medium vs.\ high $\rho = 0.96$). In other words, the L1 groups that the model scored relatively higher in the low band were also scored relatively higher in the medium and high bands.

\begin{figure}[ht]
\centering
\includegraphics[width=\linewidth]{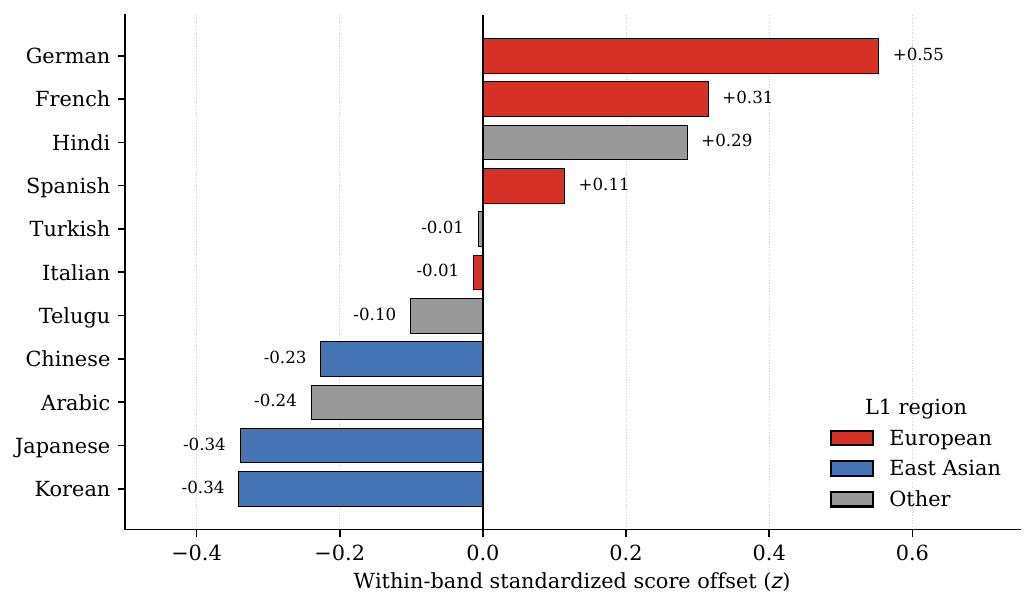}
\caption{Within-band standardized scoring offset by first language, ordered from most positive to most negative. Positive values indicate that the model assigned essays from that L1 higher raw scores than the average for their ETS band; negative values indicate lower-than-average scoring. Bars are coloured by L1 region. The European versus East-Asian separation is clearly visible: all four European-language groups have positive or near-zero offsets, while all three East-Asian-language groups have the most negative offsets.}
\label{fig:l1_offset}
\end{figure}

This offset has a clear regional structure. Grouping the L1s, essays from European-language backgrounds (German, French, Italian, Spanish) received systematically higher raw scores than essays from East-Asian-language backgrounds (Japanese, Korean, Chinese) within every band. The European--East-Asian gap was +0.21 points in the low band, +0.33 in the medium band, and +0.30 in the high band (Mann--Whitney tests, all $p < .001$). Expressed as a within-band standardized score (averaged across each L1's essays), the offset ranged from $-0.34$ (Korean) and $-0.34$ (Japanese) at one extreme to $+0.55$ (German) and $+0.32$ (French) at the other (Table~\ref{tab:offset}, Figure~\ref{fig:l1_offset}).

A concrete illustration of what this offset means in practice is provided by the low-band essays. Of the 321 essays from European-language writers rated low by ETS, the model classified 155 (48.3\%) as low, 164 (51.1\%) as medium, and 2 (0.6\%) as high; the model therefore disagreed with the ETS band for a majority (51.7\%) of these essays, and almost every disagreement (164 of 166, or 98.8\%) took the form of pushing the essay up to the medium band. Of the 500 essays from East-Asian-language writers rated low by ETS, by contrast, the model classified 330 (66.0\%) as low, 170 (34.0\%) as medium, and none as high. The roughly 18-point difference in agreement rate between the two regional groups on essays that ETS assigned to the same band is a direct manifestation of the offset. At the boundary between the low and medium bands, the model tends to keep East-Asian essays down and push European essays up.

\begin{table}[ht]
\centering
\caption{Within-band standardized Gemma score offset by L1, computed as the mean of the within-band standardized scores ($z$-scores calculated within each ETS band) across all 1,100 essays from each L1 group. Positive values indicate that the model assigned essays from that L1 higher raw scores than the average for their ETS band; negative values indicate lower-than-average scoring. Training exposure is the number of essays from each L1 in the 480-essay fine-tuning set; the 126 training essays from L1 backgrounds not represented in TOEFL11 are not shown.}
\label{tab:offset}
\begin{tabular}{lcc}
\toprule
\textbf{L1} & \textbf{Within-band score offset ($z$)} & \textbf{Training exposure ($n$)} \\
\midrule
German   & $+0.55$ & 26 \\
French   & $+0.32$ & 23 \\
Hindi    & $+0.29$ & 13 \\
Spanish  & $+0.11$ & 52 \\
Turkish  & $-0.01$ & 9 \\
Italian  & $-0.01$ & 5 \\
Telugu   & $-0.10$ & 7 \\
Chinese  & $-0.23$ & 83 \\
Arabic   & $-0.24$ & 30 \\
Japanese & $-0.34$ & 50 \\
Korean   & $-0.34$ & 56 \\
\bottomrule
\end{tabular}
\end{table}

\subsubsection{Ruling out training exposure}

We examined whether this offset could be explained by the composition of the model's fine-tuning data. The 480 training essays were unevenly distributed across L1 backgrounds (Table~\ref{tab:offset}, final column): Chinese (83), Korean (56), Spanish (52), and Japanese (50) were the most represented, while Italian (5), Telugu (7), and Turkish (9) were the least. The 11 TOEFL11 languages together account for 354 of the 480 training essays; the remaining 126 essays were written by speakers of 41 other L1s not represented in TOEFL11 (most commonly Portuguese, Russian, Tagalog, and Vietnamese) and are therefore not shown in Table~\ref{tab:offset}. If training exposure drove the offset, more heavily represented L1 groups should be scored differently in a systematic way. However, across the 11 L1 groups, the correlation between training exposure and the within-band offset was negative rather than positive, and not significant ($r = -0.43$, $p = .183$, $n = 11$); the direction is the opposite of what a training-exposure account predicts, and the pattern is directly contradicted by individual cases: Korean and Spanish had nearly identical training exposure (56 and 52 essays) but opposite offsets ($-0.34$ vs.\ $+0.11$), and Chinese, the most heavily represented L1 of all (83 essays), received one of the more negative offsets ($-0.23$). Training-set imbalance therefore does not account for the offset.

\subsubsection{The two extreme disagreements}

As noted in Section~\ref{sec:overall_performance}, only two of the 12,100 essays were misclassified by two full bands; in both cases the essay was rated low by ETS but high by the model. Because these are the model's largest disagreements with the human ground truth, we inspected both directly. Both essays happen to be from French L1 writers (out of 63 French low-band essays), responding to prompts P5 and P6, and the model assigned them raw scores of 4.0 and 4.5 respectively.

Read on their face, neither essay resembles a typical low-band response. The first presents a clear thesis on declining future car use and develops it through a balanced two-part structure (an economic argument from oil scarcity and an environmental argument from pollution), using appropriate discourse connectors and largely idiomatic English marred only by occasional spelling errors. The second is more developed still, a five-paragraph essay advancing four distinct reasons for traveling with a tour guide (safety, cultural insight, reduced stress, and social connection), each supported by a concrete personal example. At 442 words it is the longest low-band essay from any French writer in the corpus, against a low-band median of 201 words; the first essay, at 221 words, also exceeds that median. Many of the surface errors in the second essay stem from a single recurring character-encoding artifact in which an apostrophe is rendered as another symbol (e.g., in \textit{don't} and \textit{country's}).

A decisive feature of both essays is that their text is truncated mid-sentence with the first ending abruptly partway through a closing word, and the second stopping in the middle of its final clause. This truncation is the most plausible source of the disagreement. If the human raters scored complete responses while the corpus preserves only a fragment, or if the responses were incomplete or otherwise compromised in ways not visible in the retained text, a low human rating could be entirely warranted even though the visible portion reads as proficient writing. We therefore cannot determine from the available data whether these two cases reflect incomplete essay text, a labeling artifact, or a genuine human rating that the visible text does not explain. What can be said is that the model's two largest errors against the ground truth are not instances of the model misreading clearly low-quality writing as high-quality; on the visible evidence, the model's high scores are defensible, and the source of the discrepancy lies in factors outside the text it was given.

\section{\MakeUppercase{Discussion}}

This study evaluated a fine-tuned open-weight LLM (Gemma 3), trained on 480 essays from two prompts, on the full TOEFL11 corpus of 12,100 essays spanning eight unseen prompts and 11 first-language backgrounds. We discuss the findings for each research question in turn.

\subsection{Cross-prompt generalization}

The model generalized to unseen prompts with little measurable degradation. Performance was stable across all eight TOEFL11 prompts (QWK 0.677--0.730), and prompt had no significant effect on classification accuracy. This is a notable result given that none of the eight evaluation prompts appeared in the model's training data, and that the model was fine-tuned on a deliberately narrow set of only two prompts. It supports the observation by \cite{wang2024effectiveness} that fine-tuned models for this task do not appear to require a large variety of essay prompts to generalize, and extends that observation to a stricter test: here the training and evaluation prompts were entirely disjoint, whereas earlier work evaluated generalization within or adjacent to the training prompts.

The absence of any thematic-proximity advantage reinforces this interpretation. Two test prompts were judged thematically adjacent to the training prompts, yet they were scored no more accurately than the six thematically distant prompts. Taken together, these results suggest that the LoRA adapter did not learn prompt-specific scoring tied to the topics it was trained on, but rather a more general mapping from essay features to rubric-aligned scores. For practitioners, this is an encouraging finding as it implies that a scoring model of this kind can be deployed across a range of prompts without prompt-specific retraining, at least within the genre of writing represented by the TOEFL Independent Writing task.

The overall agreement of 77.79\% and QWK of 0.702 on the TOEFL11 corpus are lower than the figures reported by \cite{gayed2026aiawe} for the same model on its original held-out set (QWK 0.828, agreement 90.56\% within $\pm$0.5). This decline is expected and reflects the greater difficulty of the present evaluation rather than a failure of generalization. The TOEFL11 evaluation differs in an important way: the comparison is conducted at the band level rather than on raw scores. Since the band comparison collapses the model's fine-grained scores into three categories, an essay scored 2.5 when the true score warrants 2.0 is counted as a medium-band error even though the raw deviation is small. The near-perfect adjacent agreement (99.98\%) confirms that the model's errors remain small in magnitude. At the same time, the band-level QWK of 0.702 obtained here compares favorably with zero-shot LLM scoring of the same corpus. \cite{mizumoto2023exploring} reported a QWK of 0.388 against the same benchmark levels for GPT-3 alone, and 0.605 when GPT scores were combined with linguistic features. The only two essays misclassified by two full bands both appear, on inspection, to be cases where the visible (truncated) essay text reads as clearly proficient writing despite a low human rating (Section~4.3.4), suggesting that even the model's most extreme disagreements are not instances of it misjudging plainly low-quality writing.

\subsection{First-language effects}

The central finding of this study is that the model exhibits a systematic, L1-linked scoring offset. Within every proficiency band, essays from European-language backgrounds received higher raw scores than essays from East-Asian-language backgrounds, and this ordering was remarkably consistent across the three bands ($\rho$ = 0.88--0.96). Because the offset persists within bands, where the ETS-assigned proficiency level is approximately held constant, it cannot be attributed solely to the well-documented overall proficiency differences between these test-taker populations.

Using the available data, we were able to rule out one possible candidate explanation: the offset is not an artifact of the fine-tuning data's L1 composition. Although the 480 training essays were unevenly distributed across L1 backgrounds, training exposure did not predict the offset, and the clearest counterexample is decisive: Korean and Spanish were represented almost equally in training (56 and 52 essays) yet received opposite offsets, while Chinese, the most represented L1, received a negative offset.

What remains is that the offset is tied to the linguistic characteristics of the L1 groups themselves. Two interpretations are consistent with the evidence, and the present data cannot fully distinguish them. The first is that the offset reflects genuine within-band variation in writing quality that the coarse three-band ETS labels cannot capture: if European-background writers within the high band tend to cluster nearer the top of that band (a true score of 5.0) while East-Asian-background writers cluster nearer its floor (4.0), then the model's higher scores for the former would partly reflect a real signal. The second interpretation is that the model responds to L1-linked surface features of the writing, such as rhetorical organization, sentence structure, or characteristic error patterns, in ways that diverge from the construct the ETS rubric is intended to measure, which would constitute a genuine scoring bias. The truth may involve both mechanisms.

This finding aligns with and extends the work of \cite{liu2025enhancing}, who assessed the fairness of a fine-tuned GPT-3.5 model across four TOEFL11 L1 groups. The convergence between the two studies is notable as their model overrated German L1 essays most strongly of the four groups they examined (standardized mean difference of $+0.29$, with all four low-level German essays in their test set pushed up to the medium level), and German likewise shows the largest positive within-band offset in the present study ($+0.55$). That two independently trained models from different families (a proprietary fine-tuned GPT-3.5 and an open-weight LoRA-adapted Gemma), evaluated under different designs, shift German L1 essays upward on the same corpus suggests the effect may be a property of L1-linked features of the writing rather than an idiosyncrasy of either model. The present study extends their analysis in scope and design as it covers all 11 TOEFL11 L1 groups rather than four, evaluates 12,100 essays rather than an 880-essay test set, uses a model whose training data is fully disjoint from the evaluation corpus, and stratifies the analysis by proficiency band, thereby removing the band-composition confound that \cite{liu2025enhancing} themselves identified in their group-level comparisons. It also characterizes the effect more precisely as a directional and consistent scoring offset rather than simply uneven accuracy. It is worth emphasizing that the same underlying offset has different consequences in different bands: because the model shifts East-Asian writers' scores downward, it is more accurate on East-Asian low-band essays (which should be scored low) but less accurate on East-Asian high-band essays (which it pushes down into the medium band). The low band provides a particularly clean illustration: the model agreed with the ETS band on 66.0\% of East-Asian low-band essays but only 48.3\% of European low-band essays, and virtually every European disagreement (164 of 166) consisted of pushing the essay up to the medium band. A single scoring tendency thus manifests as both higher and lower accuracy depending on the band, which is why band-stratified analysis was essential to uncovering it.

\subsection{Implications}

These findings carry implications for the deployment of LLM-based scoring systems. The strong cross-prompt generalization suggests that such systems can be applied to new prompts without retraining, lowering the practical barrier to deployment. However, the L1-linked scoring offset means that aggregate accuracy statistics can conceal systematic differences in how a model treats writers from different language backgrounds. Any operational use of such a system, particularly in contexts approaching high-stakes assessment, should include an L1-disaggregated fairness audit rather than relying on overall agreement figures alone. The fact that the offset is consistent and directional, rather than random noise, means it could in principle be measured and, if judged to reflect bias rather than signal, corrected through calibration. Resolving whether the offset reflects genuine proficiency differences or model bias is a precondition for any such correction, and the next section outlines what that resolution would require.

\subsection{Limitations}

The present study has several limitations that constrain the interpretation of its findings. First, the TOEFL11 corpus provides only band-level scores (low, medium, high), not the raw averaged scores assigned by ETS raters. This means that evaluation metrics are necessarily coarser than those reported in \cite{gayed2026aiawe}, where raw scores were available. Metrics such as RMSE, Pearson correlation, and intraclass correlation cannot be computed against the TOEFL11 ground truth. The model's raw scores are analyzed descriptively but can only be compared against ETS at the band level. This limitation bears directly on the central finding. Since the band labels are coarse, we cannot determine whether the L1-linked scoring offset reflects genuine within-band variation in writing quality or a divergence of the model from the ETS construct. Resolving this would require access to the raw ETS scores for these essays, which would permit a direct comparison of the model's within-band ordering against the human raters' within-band ordering, or an independent second rating of a sample of essays. Neither is available within the public TOEFL11 corpus.

Second, the natural imbalance in band composition across L1 groups in the TOEFL11 corpus, while reflecting genuine population differences, complicates the L1 analysis. Some L1 $\times$ band cells contain very few essays (e.g., German low = 15, Hindi low = 29), limiting the reliability of accuracy estimates in those cells. The band-stratified analysis mitigates the confounding of overall accuracy by band composition, but the small cells at the extremes should be interpreted with caution.

Third, this study evaluates a single model (Gemma-3-27B-it), so we cannot determine from our own data whether the L1 patterns observed here are model-specific or a general property of LLM-based scoring. This scope was deliberate as the Gemma configuration was carried forward because it was the best-performing model in \cite{gayed2026aiawe}, and a sound multi-model comparison would require replicating that study's full fine-tuning and validation protocol for each additional candidate (for example, a LoRA-adapted Qwen) before subjecting it to the same 12,100-essay evaluation, which was beyond the resources available for the present study. The convergence between our German L1 results and those reported by \cite{liu2025enhancing} for a fine-tuned GPT-3.5 (Section~5.2) provides preliminary evidence that at least part of the offset may not be specific to this model, but a systematic multi-model comparison on the full corpus remains an open task, one made tractable by the public availability of the adapter and pipeline.

Fourth, the training data consists of 480 essays from two prompts, neither of which appears in the TOEFL11 corpus. While this design provides a clean test of cross-prompt generalization, it also means that the model's exposure to the diversity of argumentative topics represented in the TOEFL11 corpus is necessarily limited. Performance might improve with a more diverse training set.

Finally, this study does not examine the qualitative characteristics of misclassified essays. Understanding \textit{why} the model misclassifies certain essays, for example, whether specific L1-related writing features (rhetorical structure, error patterns, lexical choices) are associated with misclassifications, would require a follow-up qualitative analysis.

\section{\MakeUppercase{Conclusion}}

This study evaluated a fine-tuned open-weight large language model on the full TOEFL11 corpus of 12,100 essays spanning eight unseen prompts and 11 first-language backgrounds. Three findings emerged. First, the model classified essays into the three ETS proficiency bands with good performance. Second, the model generalized to unseen prompts with no significant variation in accuracy across the eight prompts and no advantage for prompts thematically related to its training data, suggesting that it learned a prompt-general scoring function rather than topic-specific artifacts. Third, and most importantly, the model exhibited a systematic, L1-linked scoring offset within every proficiency band. It assigned higher scores to writers from European-language backgrounds and lower scores to writers from East-Asian-language backgrounds, in a pattern that was highly consistent across bands and that could not be explained by the composition of the fine-tuning data.

The primary contribution of this work is the first large-scale, all-L1 fairness analysis of a fine-tuned open-weight LLM for automated essay scoring. The practical message is open-weight models adapted with modest training data generalize robustly across prompts, lowering the barrier to deployment of self-hosted, auditable scoring systems. However, aggregate accuracy can mask systematic differences in how such models score writers from different language backgrounds, and these differences should be measured directly through L1-disaggregated audits.

Several directions follow from this work. The most pressing is to determine whether the observed L1 offset reflects genuine within-band proficiency differences or a model bias, which would require either the raw ETS scores or an independent second rating. Beyond this, a qualitative analysis of misclassified essays could identify the specific L1-linked writing features that the model responds to, and a comparison across multiple models and fine-tuning configurations would establish whether the offset is specific to this model or a general property of LLM-based scoring. The open-source nature of the pipeline evaluated here makes each of these extensions tractable for the wider research community.

\section*{\MakeUppercase{Acknowledgement(s)}}
The author would like to thank the University of Pennsylvania and the Linguistic Data Consortium (LDC) for providing access to the TOEFL11 dataset used in this study.

\section*{\MakeUppercase{AI use statement}}
During the preparation of this work, the author(s) used Anthropic Claude to improve readability and cross-check claims and analysis. The author takes full responsibility for the accuracy, integrity, and originality of the final manuscript.

\section*{\MakeUppercase{Funding}}
This work is supported by the Japan Society for the Promotion of Science (JSPS) via the Grants-in-Aid for Scientific Research (Kakenhi) Grant Number 22K00718.

\bibliographystyle{apacite}
\bibliography{ourbib}

\appendix

\section{TOEFL11 evaluation prompts}\label{appendix:prompts}

The eight argumentative prompts in the TOEFL11 corpus are listed below. None overlaps with the two training prompts (TP-A, TP-B) described in Section~3.1.

\subsection*{Prompt 1 (P1):}
Do you agree or disagree with the following statement?
It is better to have broad knowledge of many academic subjects than to specialize in one specific subject.
Use specific reasons and examples to support your answer.

\subsection*{Prompt 2 (P2):}
Do you agree or disagree with the following statement?
Young people enjoy life more than older people do.
Use specific reasons and examples to support your answer.

\subsection*{Prompt 3 (P3):}
Do you agree or disagree with the following statement?
Young people nowadays do not give enough time to helping their communities.
Use specific reasons and examples to support your answer.

\subsection*{Prompt 4 (P4):}
Do you agree or disagree with the following statement?
Most advertisements make products seem much better than they really are.
Use specific reasons and examples to support your answer.

\subsection*{Prompt 5 (P5):}
Do you agree or disagree with the following statement?
In twenty years, there will be fewer cars in use than there are today.
Use reasons and examples to support your answer.

\subsection*{Prompt 6 (P6):}
Do you agree or disagree with the following statement?
The best way to travel is in a group led by a tour guide.
Use reasons and examples to support your answer.

\subsection*{Prompt 7 (P7):}
Do you agree or disagree with the following statement?
It is more important for students to understand ideas and concepts than it is for them to learn facts.
Use reasons and examples to support your answer.

\subsection*{Prompt 8 (P8):}
Do you agree or disagree with the following statement?
Successful people try new things and take risks rather than only doing what they already know how to do well.
Use reasons and examples to support your answer.


\end{document}